\documentclass[letterpaper, 10 pt, conference]{ieeeconf}  

\IEEEoverridecommandlockouts                              

\usepackage{graphicx} 
\usepackage{mathptmx} 
\usepackage{amsmath} 
\usepackage{amsfonts}
\usepackage{algorithm}
\usepackage{algorithmic}
\usepackage{subcaption}
\usepackage{float}
\usepackage{amssymb}  
\usepackage{xcolor}
\usepackage{lmodern}
\definecolor{marc}{RGB}{255, 0, 0}
\definecolor{anass}{HTML}{3bd6c6}
\definecolor{fadi}{HTML}{92268F}

\title{\LARGE \bf
 Exploring Latent Pathways: Enhancing the Interpretability of Autonomous Driving with a Variational Autoencoder
}

\author{Anass Bairouk$^{1}$, Mirjana Maras$^{1}$, Simon Herlin$^{1}$, Alexander Amini$^{2
}$, Marc Blanchon$^{1}$,\\ Ramin Hasani$^{2}$, Patrick Chareyre$^{1}$, Daniela Rus$^{2}$
\thanks{This work was supported by Capgemini Engineering.}
\thanks{$^{1}$Hybrid Intelligence part of Capgemini Engineering
        {\tt\small \it{\{first\_name.last\_name\}}@capgemini.com}}%
\thanks{$^{2}$Computer Science and Artiﬁcial Intelligence Lab, Massachusetts Institute of Technology
        {\tt\small \it{\{{amini,hasani,rus}\}}@mit.edu}}%
}

\begin{document}

\maketitle
\thispagestyle{empty}
\pagestyle{empty}

\begin{abstract}

Autonomous driving presents a complex challenge, which is usually addressed with artificial intelligence models that are end-to-end or modular in nature. Within the landscape of modular approaches, a bio-inspired neural circuit policy model has emerged as an innovative control module, offering a compact and inherently interpretable system to infer a steering wheel command from abstract visual features. Here, we take a leap forward by integrating a variational autoencoder with the neural circuit policy controller, forming a solution that directly generates steering commands from input camera images. By substituting the traditional convolutional neural network approach to feature extraction with a variational autoencoder, we enhance the system's interpretability, enabling a more transparent and understandable decision-making process.

In addition to the architectural shift toward a variational autoencoder, this study introduces the automatic latent perturbation tool, a novel contribution designed to probe and elucidate the latent features within the variational autoencoder. The automatic latent perturbation tool automates the interpretability process, offering granular insights into how specific latent variables influence the overall model's behavior. Through a series of numerical experiments, we demonstrate the interpretative power of the variational autoencoder-neural circuit policy model and the utility of the automatic latent perturbation tool in making the inner workings of autonomous driving systems more transparent.


\end{abstract}

\section{INTRODUCTION}


\begin{figure}
\centering
\includegraphics[width=0.90\columnwidth]{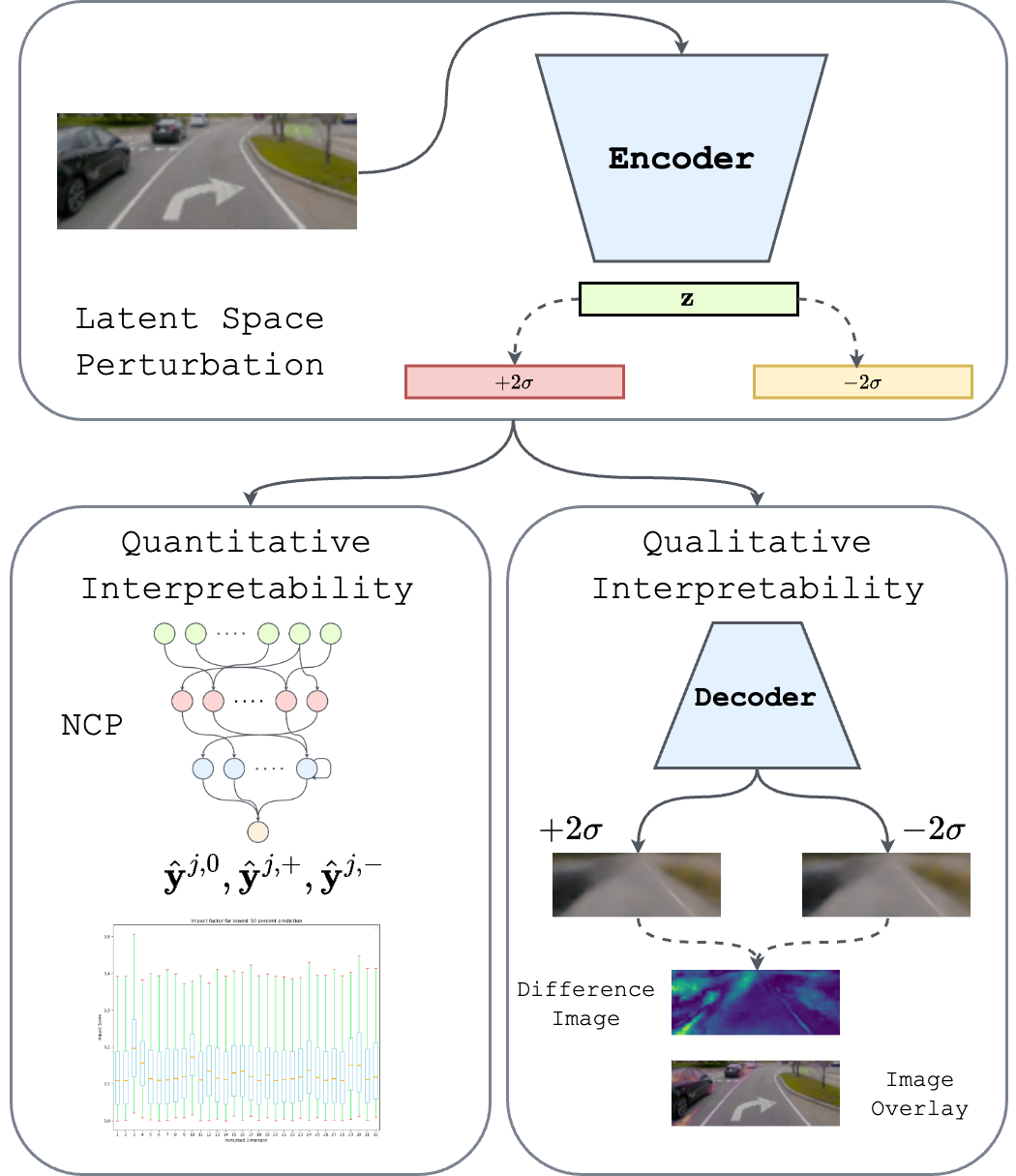}
\caption{A framework where a variational autoencoder and neural circuit policies work together, enhancing interpretability in autonomous driving from input to interpretable steering and visual outputs.}
\end{figure}

The progressive wide-spread adoption of artificial intelligence models in the domain of safety critical systems, such as autonomous driving, necessitates the development of modeling approaches with increasing levels of interpretability. Classical approaches to autonomous driving employ modular solutions, in which different functions such as perception, localization, planning and control are each modeled by a dedicated subsystem, which with the exception of perception classically rely on rule-based models \cite{chen2023end}. The rule-based approaches offer interpretability, but suffer from low generalizability to diverse conditions under which the system must operate safely. On the other hand, end-to-end learning based approaches offer high performance under diverse conditions, but are difficult to interpret due to the large model parameter space, complex parameter interactions, and lack of modularity.

In order to bridge the gap between performance and interpretability, modular learning based approaches have been developed for autonomous driving. The separate optimization of each task-specific module allows for more flexibility when explaining and probing the behavior of the whole system. When the individual machine learning based modules are in turn made more interpretable, the overall system becomes more interpretable as a whole. In the evolving landscape of modular learning based approaches, control modules implemented as a Neural Circuit Policy (NCP) within the liquid networks class of models have emerged as a significant breakthrough \cite{lechner2020neural}. This bio-inspired compact neural network receives abstract visual features from a Convolutional Neural Network (CNN) perception module, and outputs accurate and interpretable real-time steering commands. While as the outputs of a CNN can be approximately explained with posthoc explainable AI techniques, we posit that the whole system can be rendered more intrinsically explainable by replacing the CNN perception module with an inherently more explainable model. 

This paper proposes a novel approach to address the need for more interpretable modular learning based autonomous systems. Instead of relying on the conventional use of CNNs for feature extraction, here, the visual scene understanding is achieved through the lens of a Variational Autoencoder (VAE), with computed visual features then passed onto the NCP-based control module. The substitution of CNNs with VAEs aims to leverage the modular nature of VAEs, optimizing both intermediate and final outputs to significantly improve the system's interpretability. This enhancement enables a more transparent and understandable decision-making process, crucial for the safety and reliability of autonomous driving systems.

Drawing on the insights from the work of Tampuu et al. \cite{Tampuu2020EndToEndDriving}, where they underscored the importance of modular approaches in increasing interpretability within autonomous driving systems, we propose using a VAE as a theoretical optimal compressor to feed the NCP modular decision model for the inference of robust steering wheel commands. \cite{Tampuu2020EndToEndDriving} highlights the challenges posed by the lack of intermediate outputs in traditional models, which complicates the task of diagnosing why a model might misbehave. By adopting a modular architecture that optimizes both the intermediate representations through VAEs and the final decision-making output via NCPs, this paper offers a solution that enhances both interpretability and reliability.

This paper contributes: 

\begin{itemize}
    \item VAE-NCP autonomous steering solution: This solution combines a VAE perception module with an NCP control module to create a compact, robust architecture capable of generating steering commands from images. It emphasizes the optimization of both data reconstruction and decision-making processes, using a combined loss function to fine-tune model training. The architecture benefits from sparsely connected liquid time constant cells for efficient modeling and interpretable decision-making.
    \item Automatic Latent Perturbation (ALP) assistant: A novel method for VAE latent interpretability, the ALP assistant automates latent perturbation analysis. It facilitates the understanding of how each latent dimension influences the model's decisions, enhancing the interpretability of high-dimensional latent spaces and addressing limitations of traditional perturbation methods. This contribution is particularly significant for identifying and interpreting the semantic information encoded by latent variables in complex models.
\end{itemize}

 We argue that the VAE-NCP autonomous steering system and the use of the ALP assistant for increased latent interpretability, offer an overall more interpretable autonomous driving solution, compared to a CNN-NCP approach. While as visual saliency approaches are frequently used to explain CNN outputs, the faithfulness of these explanations to the CNN model they explain has been brought into question for multiple popular techniques \cite{adebayo2018sanity}. Carefully designed experiments revealed that some visual saliency techniques compute input salient regions that informed model outputs, not based on the model's operational logic but in fact on dataset characteristics that are independent of the model's reasoning processes. This insight further motivates the shift toward using VAEs within the NCP framework, aiming to establish a more direct and reliable pathway to interpret the model's decisions.

\section{Related Works}

This section will discuss previous works related to the
two main components of this contribution.
\\
\textbf{Vision-based autonomous driving.} Although rooted in the foundations of automation and robotization, autonomous driving, specifically based on perception and vision, is a recent and innovative field. Early in the research, it was necessary to delimit the boundaries and thus define the entanglements between these fields of decision-making and those of AI and perception-based cognition \cite{pomerleau1992progress}. Much more recently, in this same field and still with a strong AI and Deep learning connotation thanks to technological advances, numerous contributions have made it possible each time to define new approaches using innovative methods enabling us to push back the limits of vision-based autonomous driving. Focusing on the importance of perception and with a broad focus on the interconnections between perception and understanding of the environment, Liu et al. \cite{liu2022vision} presented a study emphasizing the importance of vision in such a task. Muhammad et al. \cite{muhammad2022vision}, with a specific focus on semantic segmentation, have linked the plurality of semantic segmentation approaches based on recent CNN approaches to the autonomous driving task, demonstrating the real connection between cognition and understanding of the vehicle environment and the control task. Thus, these last two contributions assert that the perception, understanding and description of the environment are critical to the success of the final task of autonomous driving. One can then assert, a vehicle's ability to accurately perceive and interpret its surroundings is fundamental to its autonomy and effective navigation. However, while robust perception and task completion are theoretically sufficient, we also need to be able to explain and predict the behavior of such critical systems, to ensure their viability under real-life conditions. With this in mind, Zablocki et al. \cite{zablocki2022explainability} proposed a benchmark and highlighted the crucial importance of explaining and predicting the behavior of critical systems. In their comprehensive review, Zablocki et al. delve into the methodologies and approaches for enhancing the explainability of deep learning systems in autonomous driving, emphasizing the need for these systems to not only perform tasks reliably but also to allow for interpretability and transparency in their decision-making processes, thereby ensuring safety and trustworthiness in real-world applications. Ultimately, certain research tracks seem to converge in agreeing that vision-based systems are a plausible solution to robust autonomous driving. The recent contributions of Wang et al. \cite{Wang_2023_ICCV}, Jiang et al. \cite{Jiang_2023_ICCV}, and Wang et al. \cite{Wang_2023_CVPR}, each show, through a benchmark or an innovative method, how vision can be a strong lever for driving performance and trajectory planning. As a result, the community has largely focused on proven vision-centric methods to promote robust, resilient and usable systems with the goal of developing the cars of tomorrow based on interpretable and safe vision methods.
\\
\textbf{Interpretability applied through traceable methods.} While the vision and intrinsic performance of algorithms and methods are key to the acceptance and validation of an innovative model in the field of automatic decision making applied to vehicles, it is important to underline an increasingly crucial factor in categorizing the viability of these approaches: interpretability. Zabloki et al. \cite{zablocki2022explainability} follow exactly this logic, proposing to bring notions such as interpretability, tractability, explicability or understandability to the center of contributions, making them compete with the usual evaluation of methods based on raw performance alone. The authors survey multiple methods applicable to the ADAS domain like Omeiza et al. \cite{omeiza2021explanations} and propose new methods with the goal of bringing interpretability to the domain. Jing et al. \cite{jing2022inaction} and Paleja et al. \cite{paleja2022learning} propose several concepts to anchor interpretability and explicability by conceptualizing interconnections with the ADAS domain. The trend in autonomous driving research highlights a crucial shift toward valuing interpretability as much as traditional performance metrics. This focus is pivotal for gaining trust, ensuring acceptance, and meeting regulatory standards in autonomous driving technologies. \\
VAEs are widely used structures that combine computer vision methods with traceability and/or interpretability. In the realm of computer vision and autonomous driving, variational autoencoders have emerged as a critical tool for blending algorithmic complexity with interpretability. Their utility extends to making complex vision-based models more accessible and understandable, particularly in scenarios requiring nuanced decision-making. The instance-based interpretability of VAEs, as explored by Kong and Chaudhuri \cite{kong2021understanding}, aids in deepening our understanding of how these models function and make decisions. Similarly, the work of Ainsworth et al. \cite{ainsworth2018interpretable} contributes to interpretable machine learning by disentangling complex relationships in data using VAEs. Further, Nguyen and Martínez \cite{nguyen2020learning} demonstrate how supervised VAEs can be leveraged to learn interpretable models, and Schockaert et al. \cite{schockaert2020vae} introduce VAE-LIME, showcasing VAEs' applicability in industrial contexts for local, data-driven model interpretability.
Ultimately, Liu et al. \cite{liu2020towards} propose a vision-based method to provide a visual understanding of VAEs and all their intricacies, representing a significant step in validating the need for comprehensible methods that make sense in terms of human understandability. This method's focus on visual understanding helps bridge the gap between the technical intricacies of the VAE model and human comprehension. Driven by the same objective, our work aims to offer greater interpretability and comprehensibility of the VAE algorithm and decision-making method, by providing not only a qualitative interpretation but also a quantitative one. 

\section{Methods}


\begin{figure}[!htb]
        \centering
	\includegraphics[width=0.5\textwidth,keepaspectratio]{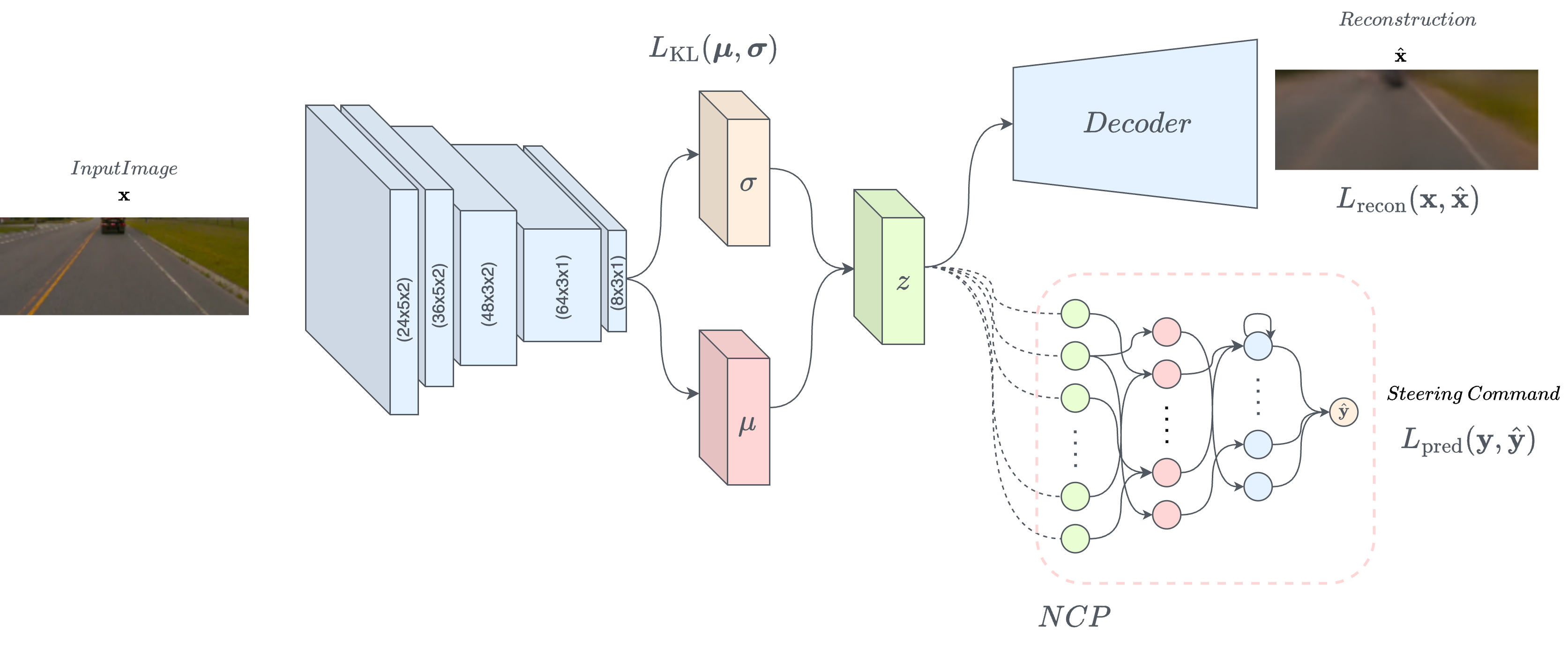}
	\caption{VAE-NCP architecture: from image to steering command and reconstruction.}
	\label{fig:e2e}
\end{figure}

In this section, the VAE, the VAE-NCP architecture and the ALP assistant are described.
\subsection{Variational autoencoder}

VAEs are valuable for creating solutions that are easy to understand. VAEs are also highly effective in compressing data, particularly useful for extracting important features in a way that makes sense to us.
They blend the domains of generative models, centered around Bayesian inference for estimating data distribution, with autoencoders, focused on encoding signals into and then decoding them from a latent space. This process is based on the assumption that the intrinsic information of a signal is contained within a manifold of lower dimensionality than the space of its representation. Without constraints on the autoencoder, interpreting the latent space can be challenging. However, VAEs hypothesize that data generation stems from a lower-dimensional probabilistic vector, with each data point drawn from its posterior distribution relative to the latent space. For computational efficiency, the latent variable is typically modeled as an independent multivariate Gaussian distribution.

The architecture of the VAE consists of an input \( \mathbf{x} \in \mathbb{R}^{h \times w \times c} \) passing through an encoder \( E \), where h, w, and c represent the height, width, and number of channels of the input image, respectively, which encodes it into a latent vector \( \mathbf{z} := E(\mathbf{x}) \hookrightarrow \mathcal{N}(\boldsymbol{\mu}, \mathbf{I}\boldsymbol{\sigma}^2) \). The latent vector is then decoded by a decoder \( D \), yielding a reconstruction \( \hat{\mathbf{x}} := D(\mathbf{z}) \). The VAE employs the loss function \( L = \beta \cdot L_{\text{recon}}(\mathbf{x}, \hat{\mathbf{x}}) + \gamma \cdot L_{\text{KL}}(\boldsymbol{\mu}, \boldsymbol{\sigma}) \), where \( L_{\text{recon}} \) is the reconstruction error and \( L_{\text{KL}} \) is the Kullback-Leibler divergence, comparing the similarity between the posterior \( p(\mathbf{z}|\mathbf{x}) \) and its prior \( p(\mathbf{z}) \). This loss function serves as a regularized reconstruction loss and promotes smooth latent representations, facilitating interpretability and robustness.


\subsection{VAE-NCP autonomous steering} 

The first contribution of this work is the VAE-NCP autonomous steering solution, as illustrated in Figure \ref{fig:e2e}. The NCP was selected for its compact neural architecture, which employs a small number of neural ordinary differential equation cells known as liquid time constants. This sparse architecture yields a robust, interpretable, and powerful framework for modeling the steering command, as demonstrated in \cite{lechner2020neural} where PCA analysis revealed that individual neurons within the NCP specialize in sub-tasks of the steering process, such as driving straight or turning.

The joint training of the VAE-NCP architecture, allows for the simultaneous optimization of both the reconstruction capability, serving dually as an optimal compressor and a feature extractor, and the decision-making aptitude of the NCP. The combined loss function \( L \) for this architecture is given by:

\begin{equation}
L = \beta \cdot L_{\text{recon}}(\mathbf{x}, \hat{\mathbf{x}}) + \gamma \cdot L_{\text{KL}}(\boldsymbol{\mu}, \boldsymbol{\sigma}) + \alpha \cdot L_{\text{pred}}(\mathbf{y}, \hat{\mathbf{y}}).
\end{equation}

The coefficients \( \beta \), \( \gamma \), and \( \alpha \) allow for the model fine-tuning and give the flexibility in choosing which loss term should be prioritized during training.

The reconstruction loss \( L_{\text{recon}} \) is quantified using the mean squared error between the input \( \mathbf{x} \) and the reconstructed image \( \hat{\mathbf{x}} \), summed over the spatial dimensions:

\begin{equation}
L_{\text{recon}} = \frac{1}{N} \sum_{i=1}^{N} \sum_{h,w} \left( \mathbf{x}_{h,w}^{(i)} - \hat{\mathbf{x}}_{h,w}^{(i)} \right)^2.
\end{equation}

The KL divergence loss \( L_{\text{KL}} \) acts as a regularization term, particularly in the context of latent space distributions. By encouraging the model's latent space to align with a predefined prior distribution, it ensures a more structured and interpretable latent space. We often choose a multivariate Gaussian as the prior for its mathematical tractability, and in this case, the KL divergence loss has a specific closed-form expression:
\begin{equation}
L_{\text{KL}} = -\frac{1}{2} \sum_{j=1}^{M} \left( 1 - \mu_j^2 - \sigma_j^2  + \log \left( \sigma_j^2 \right)  \right),
\end{equation}

where $M$ is the dimensionality of the latent vector $z$. The prediction loss \( L_{\text{pred}} \) reflects the accuracy of the steering command prediction, weighted by an exponential function of the true steering command's magnitude:

\begin{equation}
L_{\text{pred}} = \frac{\sum_{i=1}^{N} w_i \cdot \left( \hat{\mathbf{y}}^{(i)} - \mathbf{y}^{(i)} \right)^2}{\sum_{i=1}^{N} w_i}.
\end{equation}

In this equation, \(w_i = \exp \left( \lambda \cdot | \mathbf{y}^{(i)} | \right)\), with \(\lambda\) representing the factor that modulates the influence of the steering command's magnitude on the loss.

This integrative approach to training ensures that the VAE-NCP system not only accurately compresses and reconstructs input data but also effectively generates steering commands, with the latent space tailored to support both objectives.

\subsection{Automatic Latent Perturbation assistant}
\label{subsec:alp}
The second contribution of this work is the development of a new VAE latent interpretability method, which estimates the semantic information that is represented by each latent variable. 

One way to interpret the information encoded in each latent dimension of a VAE is to perform a latent perturbation analysis \cite{amini_variational_2018}. For a given input image, its latent encoding is perturbed along a single latent variable, which is then decoded to reveal the information that the latent variable accounts for. This particular approach reveals useful information about the characteristic information of each latent variable, but does not readily scale to higher dimensional latent spaces or VAEs suffering from mode collapse \cite{wang2021posterior}. A large number of latent variables cannot be efficiently analyzed with the existing perturbation approach, and non-informative latent variables cannot be readily detected.  

To remedy this problem, an automatic latent perturbation assistant was developed to automate the latent perturbation analysis and to render it directly applicable to the case of high-dimensional latent space interpretation. Given an input image $\mathbf{x}$, the tool considers each component of the corresponding latent variable ${\mathbf{z}=\left(z_j\right)_{j=1}^M}$, and generates two new reconstructions, $\hat{\mathbf{x}}^{j,+}$ and $\hat{\mathbf{x}}^{j,-}$, defined as:
\begin{equation}
    \hat{\mathbf{x}}^{j,+} = D\left(\mathbf{z} + 2 \cdot (\sigma \odot \mathbf{e}_j)\right),  
\end{equation}
\begin{equation}
        \hat{\mathbf{x}}^{j,-} = D\left(\mathbf{z} - 2 \cdot (\sigma \odot \mathbf{e}_j)\right),
\end{equation}

where $\sigma$ is the standard deviation vector, $\odot$ denotes element-wise multiplication, and $\mathbf{e}_j$ is a unit vector with the $j$-th element set to 1 and all other elements set to 0. The ALP assistant computes the difference, ${\hat{\mathbf{x}}^{j,+} - \hat{\mathbf{x}}^{j,-}}$, to approximate the input-space information that is encoded by the latent variable $z_j$. The characteristic information is computed by identifying the pixels in the difference image with values above a certain threshold, and automatically assigning a class label to each one of these pixels with the DeepLabV3+ semantic segmentation model \cite{chen2018encoder}. The classes with the majority of thresholded pixels in the difference image are then reported as the semantic information that is represented by the latent variable (see Figure \ref{fig:imagesRes}). The algorithm, as referenced in Algorithm \ref{alg:alp}, encapsulates the complete functionality of the ALP tool.

\begin{algorithm}
\caption{Auto Latent Perturbation Analysis}
\begin{algorithmic}[1]
\REQUIRE Input image $x$, Encoder $E$, Decoder $D$
\STATE $z,\ {\mu},\ {\sigma} \gets \text{E}(x)$, Encode the image to get latent space $z$ and standard deviation $\sigma$

\FOR{$j = 1$ \TO $N$}
    \STATE $\hat{\mathbf{x}}^{j,+} = D\left(\mathbf{z} + 2 \cdot (\sigma \odot \mathbf{e}_j)\right)$, perturbed reconstruction with $+2\sigma_j$
    \STATE $\hat{\mathbf{x}}^{j,-} = D\left(\mathbf{z} - 2 \cdot (\sigma \odot \mathbf{e}_j)\right)$, perturbed reconstruction with $-2\sigma_j$
    \STATE ${\Delta}\hat{\mathbf{x}}^j \leftarrow \hat{\mathbf{x}}^{j,+} - \hat{\mathbf{x}}^{j,-}$, compute the difference between the two perturbations
    \STATE ${\Delta}\hat{\mathbf{x}}^j \leftarrow {\Delta}\hat{\mathbf{x}}^j / \max({\Delta}\hat{\mathbf{x}}^j)$
    \STATE $B_j \leftarrow {\Delta}\hat{\mathbf{x}}^j > \text{quantile}({\Delta}\hat{\mathbf{x}}^j, 0.9)$, binarize difference image using the 90th percentile as the cutoff
    \STATE $R_j \leftarrow \text{label}(B_j)$, label connected components in $B_j$ to get labeled regions
    \STATE $C_j \leftarrow \text{array of zeros}$, initialize counter array for classes
    \FOR{each labeled region $l$ in $R_j$}
        \STATE $ss_{count} \leftarrow \text{count of segmentation classes in } l$
        \STATE $arg_{class} \leftarrow \text{index of max value in } ss_{count}$
        \STATE $C_j[arg_{class}] \leftarrow c[arg_{class}] + 1$, increment class counter
    \ENDFOR
\ENDFOR
\RETURN $\Delta\hat{\mathbf{x}}$, $C$
\end{algorithmic}
\label{alg:alp}
\end{algorithm}

In order to quantitatively measure and interpret the impact of each latent dimension on the model's decision-making process, particularly in the context of steering commands in autonomous driving, we further propose an extension to the automatic latent perturbation analysis. This extension involves passing the perturbations through the second head of the model, the NCP, to obtain steering error predictions. This step is crucial for understanding how variations in latent dimensions translate into changes in the model's steering decisions, providing a direct link between latent space interpretability and driving behavior.

The steering error predictions for perturbations along a given latent dimension $j$ are computed as follows:

\begin{equation}
\hat{\mathbf{y}}^{j,+} = NCP\left(\mathbf{z} + 2 \cdot (\sigma \odot \mathbf{e}_j)\right),
\end{equation}
\begin{equation}
\hat{\mathbf{y}}^{j,-} = NCP\left(\mathbf{z} - 2 \cdot (\sigma \odot \mathbf{e}_j)\right),
\end{equation}
\begin{equation}
\hat{\mathbf{y}}^{j,0} = NCP\left(\mathbf{z}\right),
\end{equation}

where $\hat{\mathbf{y}}^{j,+}$ and $\hat{\mathbf{y}}^{j,-}$ are the steering predictions for the perturbed latent vectors increased and decreased by $2\sigma$ along the $j$-th dimension, respectively, and $\hat{\mathbf{y}}^{j,0}$ is the steering prediction for the unperturbed latent vector.


The difference in predictions, which reflects the sensitivity of the steering command to perturbations in the $j$-th latent dimension, is calculated as:

\begin{equation}
\Delta \hat{\mathbf{y}}^{j,-} = \left| \hat{\mathbf{y}}^{j,-} - \hat{\mathbf{y}}^{j,0} \right|,
\end{equation}
\begin{equation}
\Delta \hat{\mathbf{y}}^{j,+} = \left| \hat{\mathbf{y}}^{j,+} - \hat{\mathbf{y}}^{j,0} \right|,
\end{equation}
\begin{equation}
\Delta \hat{\mathbf{y}}_{per} = \left| \hat{\mathbf{y}}^{j,+} - \hat{\mathbf{y}}^{j,-} \right|.
\end{equation}

To quantify the overall impact of perturbations in each latent dimension on steering decisions, we introduce the concept of an "impact score" for each dimension, denoted as $\text{I}_j$. This score aggregates the differences in steering predictions resulting from perturbations in both directions and the direct difference between these perturbed predictions, providing a comprehensive measure of the latent dimension's influence:

\begin{equation}
\text{I}_j = \frac{\Delta \hat{\mathbf{y}}^{j,-} + \Delta \hat{\mathbf{y}}^{j,+} + \Delta \hat{\mathbf{y}}_{per}}{3}.
\end{equation}

The impact score $\text{I}_j$ serves as a quantitative metric to assess the significance of each latent dimension in the model's decision-making process, particularly in steering control for autonomous driving. By evaluating $\text{I}_j$ across all latent dimensions, we can identify which dimensions have the most pronounced effect on the model's behavior, thereby highlighting the key factors that the model considers when making steering decisions. This analysis not only enhances our understanding of the model's interpretability but also guides further improvements in model design and training for better performance and reliability in autonomous driving applications.




\section{Results \& Implementation}


\subsection{Implementation}

The implementation of the VAE-NCP model was inspired by the works presented in \cite{lechner2020neural} and \cite{amini_variational_2018}. The resulting latent vector has a dimension of \(32\). The decoder mirrors the encoder architecture but utilizes transposed convolutional layers to reconstruct the input image from the latent representation. For the NCP, a recurrent neural network structure with 19 liquid time constant neurons connected with a sparsity factor of \(0.6\) is used to model the temporal dynamics of the steering command. The model receives $78\times200$ RGB images as inputs which is a dataset proposed in \cite{lechner2020neural}, comprising 5 hours of preprocessed RGB recordings, along with the corresponding driver's steering commands. 

During training, the Adam optimizer is employed with a learning rate of  $5e^{-4}$, a batch size of \(20\) and a sequence length of \(16\). The loss function weighting factors \( \beta \), \( \gamma \), and \( \alpha \) are respectively \(0.1\), \(0.001\), and \(0.066\).



During inference, the encoder and NCP components of the model work in concert to predict the steering command based on the compressed latent representation, showcasing the capability of the VAE-NCP to perform real-time decision-making tasks. 

\subsection{Offline evaluation}
\label{subsubsec:openloop}

\begin{table}
\centering
\resizebox{\columnwidth}{!}{
	\begin{tabular}{lcc}
		\hline
		\textbf{Model}                    & \textbf{Training error} & \textbf{Test error} \\ \hline
		CNN                                    & $1.41\pm0.30$                          & $4.28\pm4.63$                   \\
		CNN-RNN                        & $0.14\pm0.05$                          & $3.39\pm4.39$                   \\
		CNN-CT-GRU                              & $0.19\pm0.05$                          & $3.63\pm4.61$                    \\
		CNN-CT-RNN (19 units)             & $0.44\pm0.14$                          & $3.62\pm4.35$                    \\
		CNN-CT-RNN (64 units)             & $0.23\pm0.09$                         & $3.43\pm4.55$                    \\
		CNN-Sparse CT-RNN (19 units) & $0.77\pm0.35$                         & $4.03\pm4.80$                    \\
		CNN-Sparse CT-RNN (64 units) & $0.40\pm0.43$                        & $3.72\pm4.71$                     \\
		CNN-GRU                                       & $1.25\pm1.02$                         & $5.06\pm6.64$                    \\
		 \underline{CNN-LSTM} (64 units)                   & $\underline{\bold{0.19}}\pm0.05$                        & $\underline{\bold{3.17}}\pm3.85$          \\
		CNN-LSTM (19 units)                    & $0.16\pm0.06$                        & $3.38\pm4.48$                    \\
		CNN-Sparse LSTM (19 units)       & $1.05\pm0.57$                        & $3.68\pm5.21$                      \\
		CNN-Sparse LSTM (64 units)      & $0.29\pm0.14$                        & $3.25\pm3.93$                      \\
		CNN-NCP (19 units) \textbf{$\dagger$}                                       & $\bold{0.43}\pm0.26$                       & \textbf{$\bold{3.22}\pm3.92$  }         \\
		
  	CNN-NCP (randomly wired)         & $2.12\pm2.93$                      & $5.19\pm5.43$                        \\
		CNN-NCP (fully connected)         & $\bold{2.41}\pm3.44$                       & $\bold{5.18}\pm4.19$     \\
  \hline 
  \hline 
  VAE-LSTM (64 units) - Created Baseline              & $\bold{0.54}\pm0.26$ & $\bold{4.70}\pm4.80$            \\ 
  VAE-LSTM (19 units) - Created Baseline              & $\bold{0.60}\pm0.30$ & $\bold{6.75}\pm8.33$            \\ 
    \textit{VAE-NCP} (19 units) - Ours \textbf{$*$}               & $\bold{0.73}\pm0.22$ & $\bold{4.67}\pm3.74$            \\     
              
	\end{tabular}
        }


\caption{Comparative evaluation of passive lane-keeping performance using tenfold cross-validation. Results for CNN-based models are sourced from \cite{lechner2020neural}. We compare our model, VAE-NCP, to models with results indicated by boldface numbers, while paying special attention to three important categories: \underline{top-performing model}, top model with interpretable steering (\textbf{$\dagger$}), best end-to-end interpretable method (\textbf{$*$}).}

	\label{tab:results}
\end{table}

\begin{figure*}[!htb]
    \centering
    \includegraphics[width=0.92\textwidth,keepaspectratio]{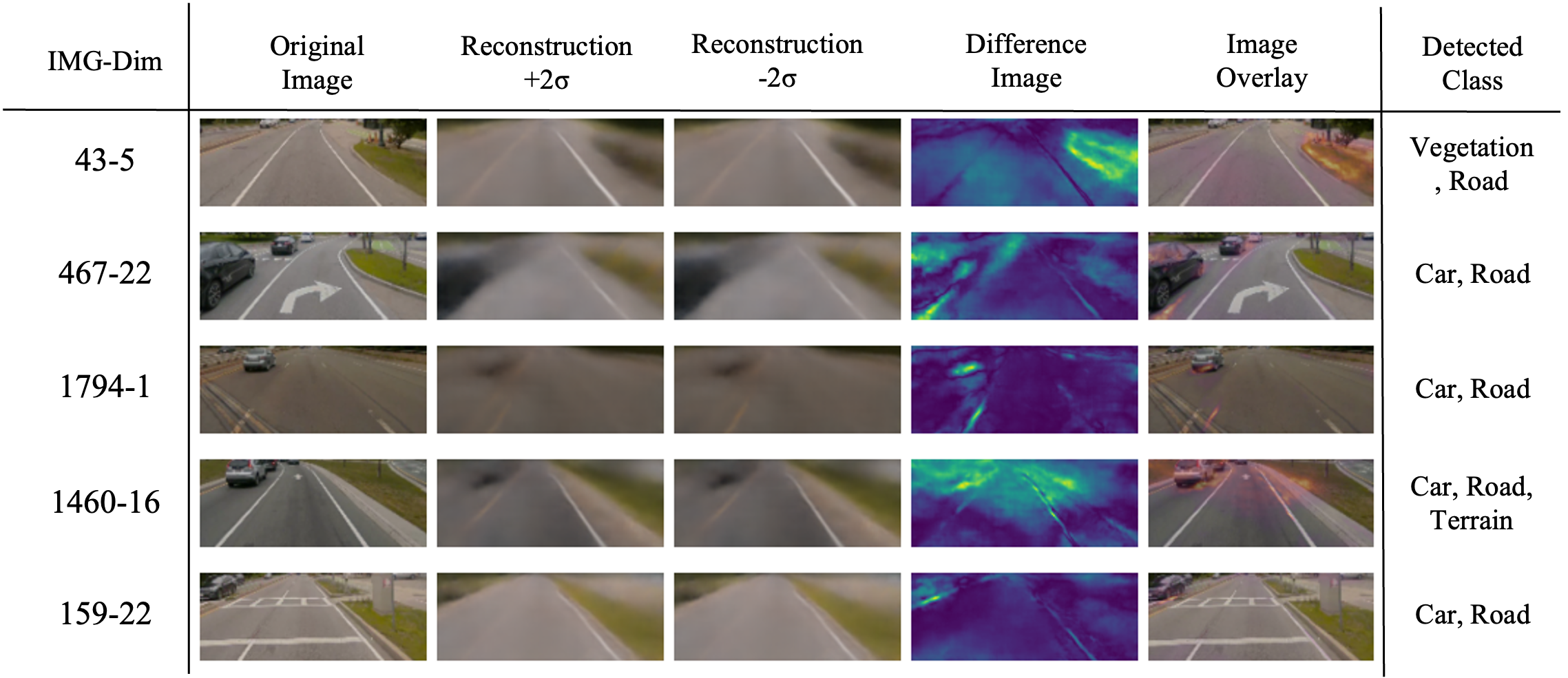}
    \caption{ALP analyses across various images and latent dimensions. Each row showcases a different image-dimension combination, the differential image accentuates all the changes,  the heatmap overlay on the original image points out the affected areas due to latent perturbations.}
    \label{fig:imagesRes}
\end{figure*}

Offline evaluations compare the behavior of an automated solution to that of a pre-recorded human driver, primarily focusing on how closely the solution mimics the driver's actions. This is particularly evident in the alignment of the solution's trajectory with the driver's. Such comparisons are made at individual time steps, which is a critical component of these evaluations. Notably, offline evaluations do not allow for the accumulation of error over time. Instead, each new time step begins with the simulation resetting to the ground-truth steering data, rather than the predicted position of the ego car by the model at that time step. This methodology ensures a clear and direct comparison of the solution's performance at each distinct moment, free from the cumulative impact of previous errors.

Additionally, to bolster the robustness of our evaluation methodology, we partitioned the dataset into ten non-intersecting segments of equal size for the purpose of tenfold cross-validation testing. This strategy, adapted from Lechner et al.~\cite{lechner2020neural}, entails training the model on nine of these segments collectively and subsequently evaluating its efficacy on the remaining segregated test segment. This process is repeated 10 times, with a different hold-out test segment in each turn. 

The results from the passive lane-keeping tenfold cross-validation evaluation (see Table \ref{tab:results}) reveal interesting insights into the performance of various models, with a special focus on the VAE-NCP model. The VAE-NCP was employed to enhance interpretability while balancing the trade-off between accuracy and interpretability. This choice was motivated by the inherent capability of VAEs to learn rich, latent representations and NCP's efficiency in parameter usage, which is significantly lower compared to traditional RNNs.

As shown in Table \ref{tab:results}, VAE-NCP is better than CNN-NCP model for the fully and randomly connected NCP, but higher compared to the best-performing models like the CNN-NCP or CNN-LSTM (64 units) with regards to the test error.
This discrepancy underscores the inherent trade-off between achieving high accuracy and enhancing model interpretability. The VAE-NCP's relatively higher test error can be attributed to its focus on generating interpretable models rather than purely optimizing for lower error rates. 

Furthermore, the use of NCP with the VAE framework illustrates a strategic move towards simplifying the model's complexity. NCPs are known for their efficiency, requiring far fewer parameters than traditional RNNs, such as GRUs and LSTMs. This efficiency does not only contribute to a leaner model that is easier to interpret but also facilitates a quicker understanding of how decisions are made within the model. Despite the increased test error, the VAE-NCP model offers a compelling advantage in applications where interpretability is key, and the decision-making process needs to be transparent, such as in autonomous driving.

\subsection{Latent Interpretability}

\begin{figure}[!htb]
	\includegraphics[width=0.5\textwidth]{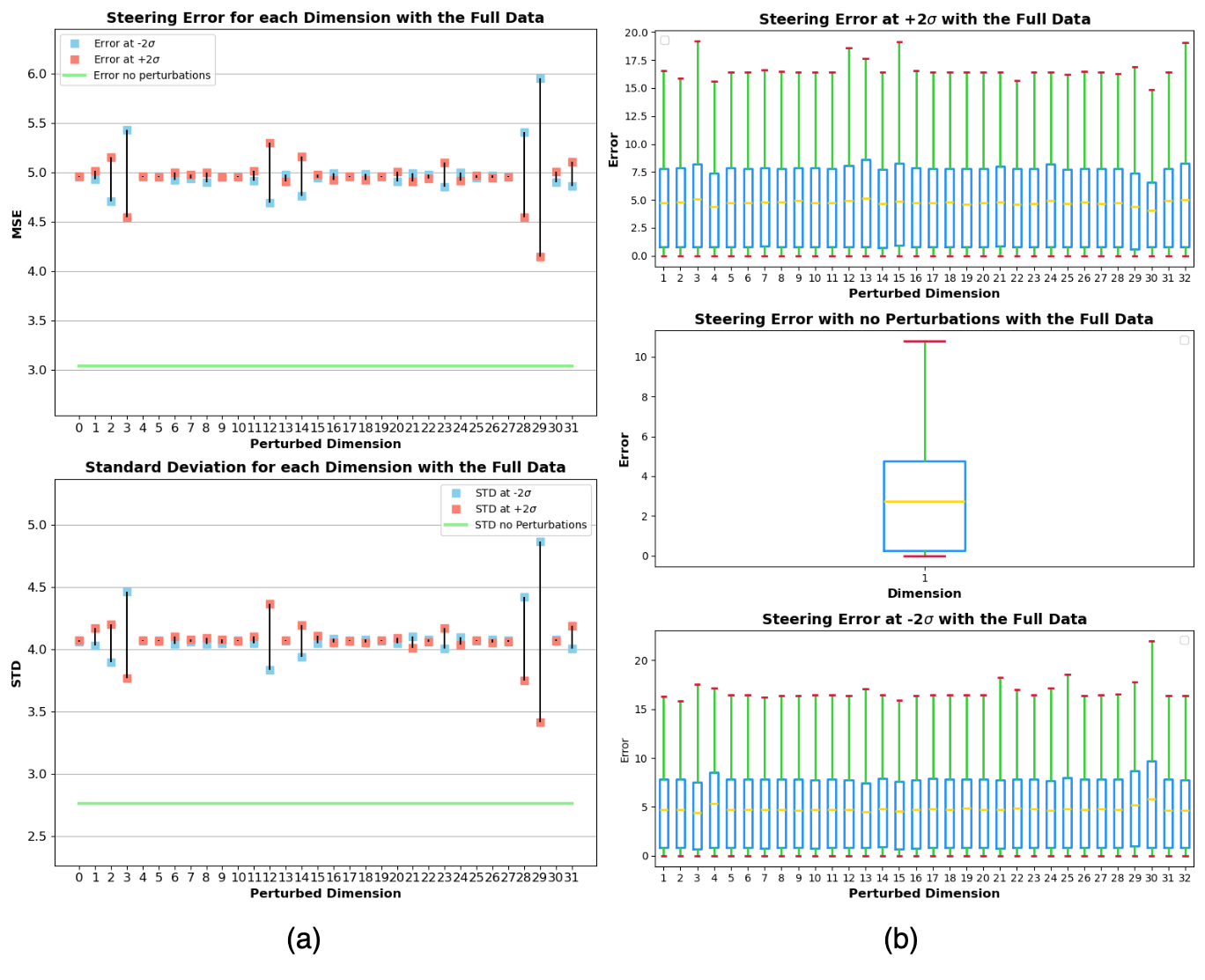}
	\caption{Analysis of steering error variability across different perturbed dimensions on a sample of 1000 images.
Top (a): the comparison of MSE at perturbations with two different standard deviations (-2$\sigma$ and +2$\sigma$) is shown for each dimension, highlighting the range of variability in the errors. Bottom (a): Standard deviation of the steering errors for each dimension. Plot (b): box plot of steering error for the whole dataset.}
	\label{fig:plotFullDataMse}
\end{figure}

In the realm of autonomous driving, interpretability is not just a luxury but a necessity for safety and reliability. The ALP tool serves as a bridge between the complexity of a VAE latent space and our understanding of its influence on model decision-making. 
Figure \ref{fig:imagesRes} offer a qualitative window into this world, where the ALP tool highlights how perturbations in specific latent dimensions affect the image reconstructions. 
In Figure \ref{fig:imagesRes}, analyzing across multiple images and latent dimensions, we gain insights into the broader interpretability of the model. Different latent dimensions are shown to be sensitive to various aspects of the input images, such as road edges or vehicle silhouettes, underscoring the dimensions' semantic relevance to the model's function.

The steering error depicted in the figure \ref{fig:plotFullDataMse} provides an initial assessment of the model's prediction variability across different latent dimensions. The top part of the figure (a) illustrates the MSE for perturbations at two standard deviations (-2$\sigma$ and +2$\sigma$),  revealing a broad range of steering error variability for each dimension. This variability is crucial for identifying which dimensions introduce the most uncertainty into the steering predictions. The bottom part of the figure (a) portrays the standard deviation of steering errors for each dimension, reinforcing the understanding of where the model's predictions fluctuate the most. Nevertheless, it's the box plot (b) that summarizes the overall distribution of steering errors, providing a macro view of the error distribution for the entire dataset. While this offers a general indication of model performance, it lacks the granularity needed to pinpoint the influence of individual latent dimensions on the model's steering decisions.

\begin{figure}[!htb]
        \centering
        \includegraphics[width=0.49\textwidth]{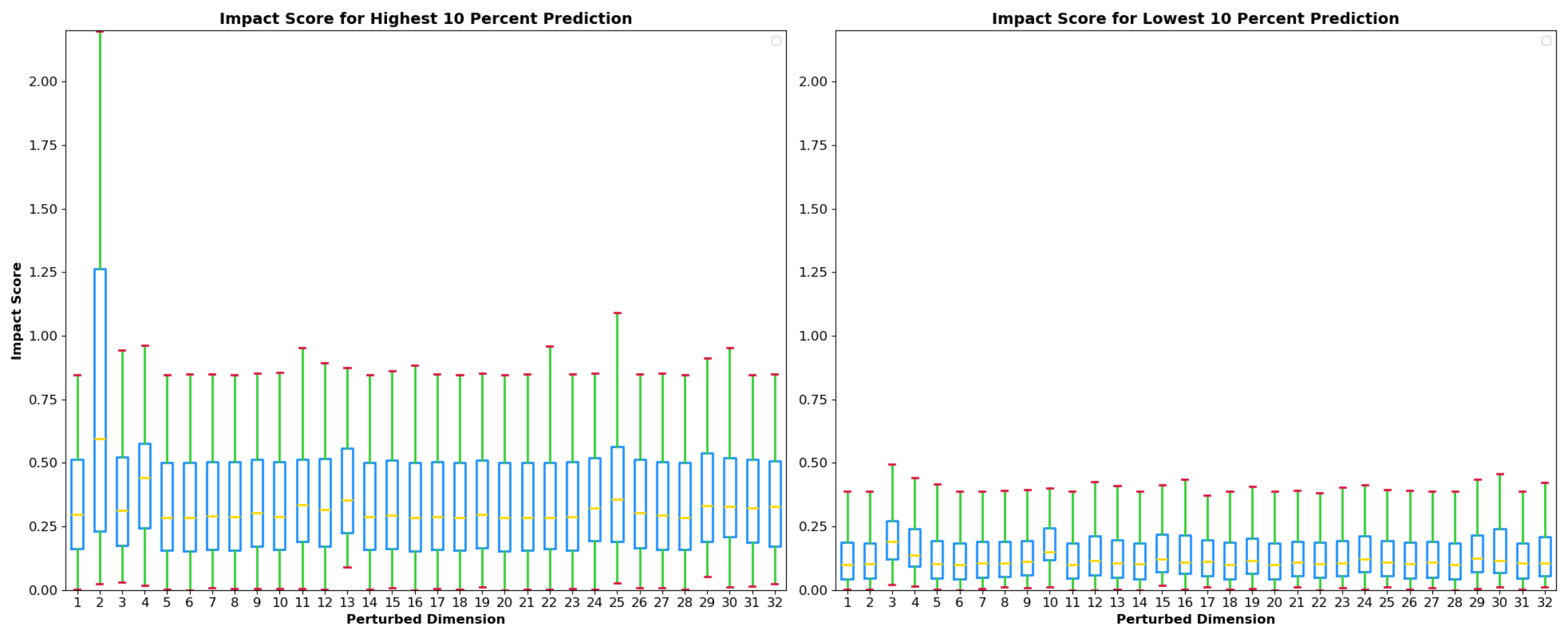}
     
	\caption{Comparison of the impact score across different latent dimensions for high and low steering prediction error percentiles. The right plot illustrates the impact scores for the highest 10\% of prediction errors, indicating dimensions with the most significant influence on steering errors. The left plot displays the impact scores for the lowest 10\% of prediction errors.} 
\label{fig:impact_score_boxplots}
\end{figure}

Analyzing the box plots in Figure \ref{fig:impact_score_boxplots} generated from the impact score, the data presents an intriguing dichotomy between the highest and lowest 10 percentiles of prediction errors. For the highest 10 percent of prediction errors, the impact scores range broadly from 0.1 to 2.5, indicating a substantial influence of certain latent dimensions on the steering predictions. Particularly, dimension 2 exhibits a pronounced variability, suggesting that perturbations along this dimension can lead to significant deviations in steering behavior. This variability can be indicative of critical features within the autonomous driving context, such as unexpected environmental variables or dynamic obstacles, that require the model to adapt its decision-making process substantially.

In stark contrast, the impact scores for the lowest 10 percent of prediction errors are tightly clustered between 0 and 0.5, reflecting a much more consistent and reliable prediction landscape. This suggests that when the model predictions are at their most accurate, the perturbations in latent dimensions do not dramatically influence the steering output, which could be attributed to more predictable and stable driving scenarios. The lower variability for dimension 2 in this percentile range further supports this, as it implies a certain robustness and reliability in scenarios where the model's performance is optimal.

These insights help us understand the model's decision-making by identifying key latent dimensions that affect prediction errors, pinpointing improvement areas. This is crucial for spotting failure modes and Out-Of-Distribution (OOD) cases, situations the model hasn't seen during training. By concentrating on latent dimensions with significant impact, especially those with high variability, we can enhance data collection and refine the model, boosting its performance and reliability in real-world autonomous driving applications.





\section{Conclusions}

The conclusion of our study underscores the nuanced balance between interpretability and accuracy within autonomous driving models, exemplified by our VAE-NCP framework. Our offline evaluation showcases the VAE-NCP model's distinct approach, prioritizing interpretability through a modular and transparent architecture, despite a comparative increase in test error. This decision reflects our commitment to enhancing safety and reliability in autonomous driving by enabling a deeper understanding of model decision-making processes. Furthermore, the application of the automated latent perturbation tool reveals critical insights into how specific latent dimensions influence steering predictions, highlighting areas for targeted improvement and model robustness against OOD scenarios. Ultimately, our findings advocate for a more interpretable and reliable framework in autonomous systems, where understanding and trust in the technology are as crucial as performance metrics.
From the insights gleaned through our research, it is clear that the  ALP tool not only enhances our understanding of model behavior in autonomous driving applications but also opens avenues for future innovations. The application of the ALP tool in active learning is particularly promising. By utilizing the impact score to identify and select challenging examples for further labeling and retraining, we can significantly improve model performance. 











\bibliographystyle{IEEEtranS}
\bibliography{Biblio.bib,biblio_sota}

\begin{thebibliography}{10}
\providecommand{\url}[1]{#1}
\csname url@samestyle\endcsname
\providecommand{\newblock}{\relax}
\providecommand{\bibinfo}[2]{#2}
\providecommand{\BIBentrySTDinterwordspacing}{\spaceskip=0pt\relax}
\providecommand{\BIBentryALTinterwordstretchfactor}{4}
\providecommand{\BIBentryALTinterwordspacing}{\spaceskip=\fontdimen2\font plus
\BIBentryALTinterwordstretchfactor\fontdimen3\font minus \fontdimen4\font\relax}
\providecommand{\BIBforeignlanguage}[2]{{%
\expandafter\ifx\csname l@#1\endcsname\relax
\typeout{** WARNING: IEEEtranS.bst: No hyphenation pattern has been}%
\typeout{** loaded for the language `#1'. Using the pattern for}%
\typeout{** the default language instead.}%
\else
\language=\csname l@#1\endcsname
\fi
#2}}
\providecommand{\BIBdecl}{\relax}
\BIBdecl

\bibitem{adebayo2018sanity}
J.~Adebayo, J.~Gilmer, M.~Muelly, I.~Goodfellow, M.~Hardt, and B.~Kim, ``Sanity checks for saliency maps,'' \emph{Advances in neural information processing systems}, vol.~31, 2018.

\bibitem{ainsworth2018interpretable}
S.~Ainsworth, N.~Foti, A.~K. Lee, and E.~Fox, ``Interpretable vaes for nonlinear group factor analysis,'' \emph{arXiv preprint arXiv:1802.06765}, 2018.

\bibitem{amini_variational_2018}
\BIBentryALTinterwordspacing
A.~Amini, W.~Schwarting, G.~Rosman, B.~Araki, S.~Karaman, and D.~Rus, ``\BIBforeignlanguage{en}{Variational {Autoencoder} for {End}-to-{End} {Control} of {Autonomous} {Driving} with {Novelty} {Detection} and {Training} {De}-biasing},'' in \emph{\BIBforeignlanguage{en}{2018 {IEEE}/{RSJ} {International} {Conference} on {Intelligent} {Robots} and {Systems} ({IROS})}}.\hskip 1em plus 0.5em minus 0.4em\relax Madrid: IEEE, Oct. 2018, pp. 568--575. [Online]. Available: \url{https://ieeexplore.ieee.org/document/8594386/}
\BIBentrySTDinterwordspacing

\bibitem{chen2023end}
L.~Chen, P.~Wu, K.~Chitta, B.~Jaeger, A.~Geiger, and H.~Li, ``End-to-end autonomous driving: Challenges and frontiers,'' \emph{arXiv preprint arXiv:2306.16927}, 2023.

\bibitem{chen2018encoder}
L.-C. Chen, Y.~Zhu, G.~Papandreou, F.~Schroff, and H.~Adam, ``Encoder-decoder with atrous separable convolution for semantic image segmentation,'' in \emph{Proceedings of the European conference on computer vision (ECCV)}, 2018, pp. 801--818.

\bibitem{Jiang_2023_ICCV}
B.~Jiang, S.~Chen, Q.~Xu, B.~Liao, J.~Chen, H.~Zhou, Q.~Zhang, W.~Liu, C.~Huang, and X.~Wang, ``Vad: Vectorized scene representation for efficient autonomous driving,'' in \emph{Proceedings of the IEEE/CVF International Conference on Computer Vision (ICCV)}, October 2023, pp. 8340--8350.

\bibitem{jing2022inaction}
T.~Jing, H.~Xia, R.~Tian, H.~Ding, X.~Luo, J.~Domeyer, R.~Sherony, and Z.~Ding, ``Inaction: Interpretable action decision making for autonomous driving,'' in \emph{European Conference on Computer Vision}.\hskip 1em plus 0.5em minus 0.4em\relax Springer, 2022, pp. 370--387.

\bibitem{kong2021understanding}
Z.~Kong and K.~Chaudhuri, ``Understanding instance-based interpretability of variational auto-encoders,'' \emph{Advances in Neural Information Processing Systems}, vol.~34, pp. 2400--2412, 2021.

\bibitem{lechner2020neural}
M.~Lechner, R.~Hasani, A.~Amini, T.~A. Henzinger, D.~Rus, and R.~Grosu, ``Neural circuit policies enabling auditable autonomy,'' \emph{Nature Machine Intelligence}, vol.~2, no.~10, pp. 642--652, 2020.

\bibitem{liu2022vision}
F.~Liu, Z.~Lu, and X.~Lin, ``Vision-based environmental perception for autonomous driving,'' \emph{Proceedings of the Institution of Mechanical Engineers, Part D: Journal of Automobile Engineering}, p. 09544070231203059, 2022.

\bibitem{liu2020towards}
W.~Liu, R.~Li, M.~Zheng, S.~Karanam, Z.~Wu, B.~Bhanu, R.~J. Radke, and O.~Camps, ``Towards visually explaining variational autoencoders,'' in \emph{Proceedings of the IEEE/CVF Conference on Computer Vision and Pattern Recognition}, 2020, pp. 8642--8651.

\bibitem{muhammad2022vision}
K.~Muhammad, T.~Hussain, H.~Ullah, J.~Del~Ser, M.~Rezaei, N.~Kumar, M.~Hijji, P.~Bellavista, and V.~H.~C. de~Albuquerque, ``Vision-based semantic segmentation in scene understanding for autonomous driving: Recent achievements, challenges, and outlooks,'' \emph{IEEE Transactions on Intelligent Transportation Systems}, 2022.

\bibitem{nguyen2020learning}
A.-p. Nguyen and M.~R. Mart{\'\i}nez, ``Learning invariances for interpretability using supervised vae,'' \emph{arXiv preprint arXiv:2007.07591}, 2020.

\bibitem{omeiza2021explanations}
D.~Omeiza, H.~Webb, M.~Jirotka, and L.~Kunze, ``Explanations in autonomous driving: A survey,'' \emph{IEEE Transactions on Intelligent Transportation Systems}, vol.~23, no.~8, pp. 10\,142--10\,162, 2021.

\bibitem{paleja2022learning}
R.~Paleja, Y.~Niu, A.~Silva, C.~Ritchie, S.~Choi, and M.~Gombolay, ``Learning interpretable, high-performing policies for autonomous driving,'' \emph{arXiv preprint arXiv:2202.02352}, 2022.

\bibitem{pomerleau1992progress}
D.~A. Pomerleau, ``Progress in neural network-based vision for autonomous robot driving,'' in \emph{Proceedings of the Intelligent Vehicles92 Symposium}.\hskip 1em plus 0.5em minus 0.4em\relax IEEE, 1992, pp. 391--396.

\bibitem{schockaert2020vae}
C.~Schockaert, V.~Macher, and A.~Schmitz, ``Vae-lime: deep generative model based approach for local data-driven model interpretability applied to the ironmaking industry,'' \emph{arXiv preprint arXiv:2007.10256}, 2020.

\bibitem{Tampuu2020EndToEndDriving}
\BIBentryALTinterwordspacing
A.~Tampuu, T.~Matiisen, M.~Semikin, D.~Fishman, and N.~Muhammad, ``A survey of end-to-end driving: Architectures and training methods,'' \emph{IEEE Transactions on Neural Networks and Learning Systems}, 2020. [Online]. Available: \url{https://arxiv.org/abs/2003.06404}
\BIBentrySTDinterwordspacing

\bibitem{Wang_2023_ICCV}
N.~Wang, Y.~Luo, T.~Sato, K.~Xu, and Q.~A. Chen, ``Does physical adversarial example really matter to autonomous driving? towards system-level effect of adversarial object evasion attack,'' in \emph{Proceedings of the IEEE/CVF International Conference on Computer Vision (ICCV)}, October 2023, pp. 4412--4423.

\bibitem{Wang_2023_CVPR}
X.~Wang, Z.~Zhu, Y.~Zhang, G.~Huang, Y.~Ye, W.~Xu, Z.~Chen, and X.~Wang, ``Are we ready for vision-centric driving streaming perception? the asap benchmark,'' in \emph{Proceedings of the IEEE/CVF Conference on Computer Vision and Pattern Recognition (CVPR)}, June 2023, pp. 9600--9610.

\bibitem{wang2021posterior}
Y.~Wang, D.~Blei, and J.~P. Cunningham, ``Posterior collapse and latent variable non-identifiability,'' \emph{Advances in Neural Information Processing Systems}, vol.~34, pp. 5443--5455, 2021.

\bibitem{zablocki2022explainability}
{\'E}.~Zablocki, H.~Ben-Younes, P.~P{\'e}rez, and M.~Cord, ``Explainability of deep vision-based autonomous driving systems: Review and challenges,'' \emph{International Journal of Computer Vision}, vol. 130, no.~10, pp. 2425--2452, 2022.

\end{thebibliography}





\end{document}